\documentclass[a4paper,twoside]{article}

\usepackage[utf8]{inputenc} 
\usepackage[T1]{fontenc} 
\usepackage{RR,RRthemes}
\usepackage{hyperref}
\usepackage{mdwlist}
\usepackage{alltt}
\usepackage{times}
\usepackage{named}
\usepackage{lscape}
\usepackage{rotating}
\usepackage{comment} 
\usepackage{pgf}
\usepackage{tikz}

\newenvironment{seconditemize}%
{ \begin{list}%
	{--}%
	{\setlength{\labelwidth}{5pt}%
	 \setlength{\leftmargin}{-2pt}%
	 \setlength{\topsep}{0pt}%
	 \setlength{\itemsep}{0pt}}}%
{ \end{list} }

\newcommand{\ontoset}{
\tikzstyle{class}=[shape=rectangle, rounded corners, minimum size=.5cm,draw,font=\sf] 
\tikzstyle{type}=[shape=rectangle,color=blue, minimum size=.3cm,draw,font=\tt] 
\tikzstyle{attr}=[color=green!50!black,minimum size=.5cm,font=\sf] 
\tikzstyle{inst}=[font=\itshape,minimum size=.5cm,draw=blue,thick,font=\sf] 
\tikzstyle{data}=[minimum size=.5cm,fill=blue!20,font=\tt] 
\tikzstyle{restr}=[shape=rectangle,minimum size=.5cm,fill=purple!20,font=\sf]
\tikzstyle{axiom}=[shape=rectangle,minimum size=.5cm,fill=orange!20,font=\it]

\tikzstyle{super}=[>=latex,->,thick]
\tikzstyle{parent}=[>=latex,-,thick]
\tikzstyle{const}=[>=latex,->,thin, densely dotted]
\tikzstyle{excl}=[snake=zigzag,thick]
\tikzstyle{isa}=[>=latex,->,thin]
\tikzstyle{val}=[>=latex,->,thin]

\tikzstyle{corresp}=[color=blue,<->,very thick]
\tikzstyle{corrbox}=[rectangle,color=blue,very thick]
}

\newcommand\smtt[1]{\texttt{\small #1}}
\newenvironment{smalltt}%
{\scriptsize\begin{alltt}{}}{\end{alltt}}

\RRdate{July 2011}

\RRauthor{
François Scharffe\thanks{LIRMM, Montpellier, France \smtt{Francois.Scharffe@lirmm.fr}; part of this work was achieved when this author was at INRIA Grenoble Rhône-Alpes.}%
  \and
Jérôme Euzenat\thanks{INRIA \& LIG, Montbonnot, France \smtt{Jerome.Euzenat@inria.fr}}%
}
\authorhead{MeLinDa}
\RRtitle{MeLinDa:\\
un cadre pour le liage des données du web}
\RRetitle{MeLinDa:\\
an interlinking framework for the web of data}
\titlehead{an interlinking framework for the web of data}
\RRnote{This work as been partially published as \protect\cite{scharffe2010a}.}
\RRresume{Le web des données consiste à publier des données sur le web de telle sorte qu'elles puissent être interprétées et connecttées entre elles.
Il est donc vital d'établir les liens entre ces données à la fois pour le web des données et pour le web sémantique qu'il contribue à nourrir.
Nous considérons les diverses techniques proposées à cette fin et analysons leurs similarités et différences.
Nous proposons un cadre général dans lequel s'inscrivent les différentes techniques utilisées et nous montrons comment elles s'y insèrent.
Ce cadre permet de considérer les relations entre le liage de données et l'alignement d'ontologies: bien que ces activités puissent être considérées comme similaires (elles trouvent les relations entre entités), elles n'ont pas le même but mais bénéficieraient à  collaborer.
Nous proposons une architecture permettant aux outils de liage de tirer parti des alignements entre ontologies.
}
\RRabstract{The web of data consists of data published on the web in such a way that they can be interpreted and connected together.
It is thus critical to establish links between these data, both for the web of data and for the semantic web that it contributes to feed. 
We consider here the various techniques developed for that purpose and analyze their commonalities and differences. 
We propose a general framework and show how the diverse techniques fit in the framework.
From this framework we consider the relation between data interlinking and ontology matching. 
Although, they can be considered similar at a certain level (they both relate formal entities), they serve different purposes, but would find a mutual benefit at collaborating. 
We thus present a scheme under which it is possible for data linking tools to take advantage of ontology alignments. 
}
\RRmotcle{Web s\'emantique, liage de données, alignement d'ontologies, web des donn\'ees}
\RRkeyword{Semantic web, data interlinking, instance matching, ontology alignment, web of data}

\RRprojets{exmo}
\RRdomaine{1} 
\RRthemeProj{exmo} 
\RRdomaineProjBis{exmo} 
\RCGrenoble 

\begin{document}

\RRNo{7691}
\makeRR   

\setcounter{page}{1}

\section{Introduction}

The web of data is the network resulting from publishing structured data sources in RDF and interlinking these data sources with explicit links. 
A large quantity of structured data is being published particularly through the Linking Open Data project\footnote{\url{http://esw.w3.org/topic/SweoIG/TaskForces/CommunityProjects/LinkingOpenData}}. 
Web data sets are expressed according to one or more vocabularies or ontologies, which range from simple database schema exposure to full-fledged ontologies.

The web of data requires to interlink the various published data sources. 
Given the large amount of published data, it is necessary to provide means to automatically link those data. 
Many tools were recently proposed in order to solve this problem, each having its own characteristics (see Section~\ref{sec:analysis}).

In many cases, data sets containing similar resources are published using different ontologies.
Hence, data interlinking tools need to reconcile these ontologies before finding the links between entities.
This could be done automatically, but more often this is done manually and built in the link specifications.
This has two drawbacks: 
(a) this prevents to reuse the work made in ontology matching for reconciling ontologies, and 
(b) the information about reconciling the ontologies is mixed with the information about how to identify entities.

Hence, the goal of this work is to analyse existing interlinking tools and to determine (1) how they fit in the same framework, (2) if it is possible to define a language for specifying the linking techniques to be used, and (3) how is data interlinking related to ontology matching.
This report contributions are as follows:
\begin{itemize*}
 \item A comprehensive survey of existing data interlinking tools,
 \item A characterization of task/problem categories for web data set interlinking,
 \item A proposal for improving data interlinking tools with ontology alignments.
\end{itemize*}

For that purpose, after briefly introducing the challenges of data interlinking and ontology matching (Section~\ref{sec:background}),
we  provide a general framework for data interlinking in which all these tools can be included (Section~\ref{sec:framework}).
From this analysis, we review six data interlinking tools and the way they are built (Section~\ref{sec:analysis}). 
This framework clearly separates the data interlinking and ontology matching activities and we show how these can collaborate through three different languages for links, data linking specifications and ontology alignments (Section~\ref{sec:coop}). 
We provide examples of an expressive alignment language (Section~\ref{sec:edoal}) and a linking specification language (Section~\ref{sec:silk}).
Finally, we show how these two languages can be adapted for cooperating (Section~\ref{sec:extsilk}).

\section{Web of data, data interlinking, and ontology alignment}\label{sec:background}

We briefly introduce linked data and the data interlinking problem.
We provide examples of this problem and why it would require specific linking tools.
We then present why these tools could take advantage of ontology matching and alignments.

\subsection{Linked data}

The web of data is based on the following four principles \cite{bernerslee:2009,heath2011a}:
\begin{enumerate*}
 \item Resources are identified by URIs.
 \item URIs are dereferenceable.
 \item When a URI is dereferenced, a description of the identified resource should be returned, ideally adapted through content negotiation.
 \item Published web data sets must contain links to other web data sets.
\end{enumerate*}

As long as they follow these rules, \emph{linked data} can be published in various ways (RDF data sets, SPARQL endpoints, XHTML+RDFa pages \cite{adida2008b}, databases exposed through HTTP  \cite{bizer:2003,sahoo2009a}).
Web data sets can also be constructed collaboratively, through the use of specialized tools \cite{volkel:2006}.

\subsection{The data interlinking problem and linksets}

\begin{figure}[!ht]
\begin{center}
\begin{tikzpicture}[cap=round]

\draw (0,3) node (u1) {\textsf{URI1}};
\draw (7,3) node (u2) {\textsf{URI2}};

\draw (3.5,2.5) node[draw,rounded corners] (di) {Data interlinking};

\draw[<->] (u1.north east) .. controls +(2cm,7.5mm) and +(-2cm,7.5mm) .. node[above] {\textsf{owl:sameAs}} (u2.north west);
\draw[->,dotted] (u1) |- (di);
\draw[->,dotted] (u2) |- (di);
\draw[->,dotted] (di) -- (3.5,3.75);

\end{tikzpicture} 
\end{center}
 \caption{The data interlinking problem.}
 \label{fig:general-schema}
\end{figure}
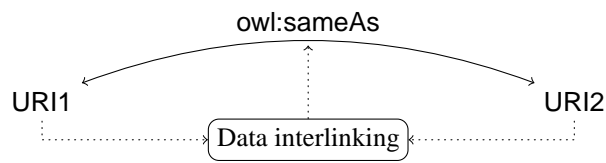

A main problem on the web of data is to create links between entities of different data sets.
Most often, this consists of identifying the same entity across different data sets and publishing a link between them as a \smtt{owl:sameAs} statement (shortened as \smtt{sameAs} hereafter).
We call this task data interlinking and summarize it in Figure~\ref{fig:general-schema}.

Once identified, the links discovered between two data sets must also be published in order to be reused. The VoiD vocabulary \cite{alexander:2009} allows for describing \emph{linksets} as special data sets containing links between resources of two given data sets. A linkset is represented as an RDF named graph described using VoiD annotations, as shown in the RDF/N3 code below:

\begin{scriptsize}
\begin{verbatim}
{
 <http://www.example.org/linkset/DBPedia-MB>
  a void:Linkset ;
  void:target <http://www.dpbedia.org>;
  void:target <http://www.musicbrainz.org>;
}
<http://www.example.org/linkset/DBPedia-MB>
{
 <http://www.dbpedia.org/resource/Johann_Sebastian_Bach>
 owl:sameAs
 <http://www.musicbrainz.org/artist/24f1766e-9635-4d58-a4d4-9413f9f98a4c> .
}
\end{verbatim}
\end{scriptsize}

Once linksets are constructed, two approaches are proposed to retrieve equivalences between resources: it is possible to assign to each real world entity a global identifier that will then be related to every URIs describing this entity. This is the approach taken in the OKKAM project \cite{bouquet:2008} that proposes the usage of Entity Name Servers taking the role of resource name repositories. The other approach uses equivalence lists maintained with interlinked resources across data sets. There is thus no global identifier in this approach but equivalence links can be followed using a third-party web service, e.g., \url{http://sameas.org}, or a bilatteral protocol \cite{volz2009a}.

The \emph{data interlinking} task can be achieved manually or with the help of data interlinking tools.
These tools take as input two data sets and ultimately provide a linkset.
In addition, they use what we call a \emph{linking specification}, i.e., a ``script'' specifying how and/or what to link. 
Indeed, given data set sizes, the search space for resources interlinking can reach several billion resources\footnote{4.2 billion RDF triples related by 142 million links: source Wikipedia, May 2009.}. 
It is thus necessary to use heuristics giving hints to the interlinking system about where to look for the corresponding resources in the two data sets. 
These linking specifications can be specific to a pair of data sets and can be reused for regenerating linksets (we provide an example of such a specification in the Silk language in Section~\ref{sec:silk}).

\subsection{Interlinking data sets}\label{sec:interlinking-datasets}

Mining for similar resources in two web data sets raises many problems. Each data set having its own namespace, resources in different data sets are given different URIs. Also, although naming conventions exist \cite{sauermann:2008}, there is no formal nor standard way of naming resources. For example, if we take the URI for the famous musician Johann Sebastian Bach in various web data sets we obtain different results though they all represent the same real world object (Table~\ref{tab:uri-variation}). 

\begin{table}[h!]
\begin{center}
{\setlength{\tabcolsep}{3pt}
\begin{tabular}{|l|l|}
\hline
Dataset & URI\\ \hline
MusicBrainz & http://musicbrainz.org/artist/24f1766e-9635-4d58-a4d4-9413f9f98a4c \\ \hline
LastFM & http://www.last.fm/music/Johann+Sebastian+Bach \\ \hline
DBpedia & http://dbpedia.org/resource/Johann\_Sebastian\_Bach \\ \hline
OpenCyc & http://sw.opencyc.org/concept/Mx4rwJw6npwpEbGdrcN5Y29ycA \\
\hline
\end{tabular}
}
\end{center}
\caption{Varying URIs across different data sets.}
\label{tab:uri-variation}
\end{table}

As this example demonstrates, URIs are different across data sets, both because of their namespaces and because of their fragments. 
Fragments are generated according to two strategies: an internal ID as for MusicBrainz and OpenCyc, or the concatenation of some of the resource properties, as for LastFM and DBpedia. 
When the first strategy is used, an interlinking system might not be able to find correspondences between two resources by looking at URIs only. 

Fortunately, dereferencing URIs can be used for retrieving more information about entities: property values and related resources can be observed.
But for the same real-world entity, the same property can take different values, making the interlinking process more difficult. 
This can be because of varying value approximations across data sets, because of different units of measure, because of mistakes in the data sets, or because of loose ontological specifications. For instance, the property \smtt{foaf:name} does not specify in what format should the name be given. ``J.S. Bach'', ``Bach, J.S.'' or ``Johann Sebastian Bach'' are possible values for this property.
Hence, data interlinking tools have to compare property values in order to decide if two entities are the same, and must be linked, or not.
For that purpose, tools use similarity measures based on the type of values, e.g., string, numbers, dates, and aggregate the results of these measures.
This activity is reminiscent of \emph{record linkage} in database which has been given considerable attention \cite{fellegi:1969,winkler:2006,elmagarmid:2007}.
The tools studied in Section~\ref{sec:analysis} reuse many of the record linkage techniques.

Another problem is caused by the usage of heterogeneous ontologies for describing data sets. 
In this case, a same resource is typed according to different classes and described with different 
predicates belonging to different ontologies. 
For example, a name in a data set can be attributed using the \smtt{foaf:name} data property from the FOAF ontology while it is attributed using the \smtt{vcard:N} object property from the VCard ontology in another data set. 

Hence, for the interlinking techniques to work, it is necessary that the data sets use the same ontology or that data interlinking tools are aware of the correspondences between ontologies.

\subsection{Ontology matching and alignment}

\emph{Ontology matching} allows for finding correspondences between ontology entities \cite{euzenat:2007b}. 
The result of this process is called an \emph{ontology alignment}. 
Once the ontologies matched, the alignment can be stored and retrieved when an application needs to use data described according to another ontology \cite{euzenat:2007c}.

Matching ontologies requires to overcome the mismatches originating from the different conceptualizations of the domains described by ontologies \cite{visser:1997,klein:2001}. These mismatches may be of different nature: terminological mismatches concern differences of naming such as the usage of synonym terms for concept labeling; conceptual mismatches concern different conceptualizations of the domain such as structuring along different properties; structural mismatches concern heterogeneous structures, like different granularities in the class hierarchies. 
Ontology matching is similar to database schema matching \cite{rahm:2001}. 
Specific works on ontology matching were proposed in the last ten years \cite{noy:2000} that now reach maturity \cite{euzenat:2007b}.
It is not the purpose of this paper to describe any particular technique.

While different URI constructions and variations of property values can find automatic solutions, the problem of having heterogeneous ontologies is in most interlinking tools solved by manually specifying the correspondences (see Table~\ref{tab:datalinkers}, Section~\ref{sec:analysis}). 
This considerably complexifies the interlinking process. 
Ontology matching techniques can be used to facilitate the interlinking task and ontology alignments reused in linking specifications.

\bigskip
The goal of this paper is to investigate the relationships between data interlinking and ontology matching. 
In particular, we want to understand if these two activities should be merged into a single activity and share the same formats or if there are good reasons to keep them separated.
In the second perspective, we also want to establish how they can benefit from one another.
For that purpose, we analyzed available systems for data interlinking.

\section{A framework for data interlinking}\label{sec:framework}

We provide, in this section, a general framework encompassing the various approaches used to interlink resources on the web of data. 

In the most general case illustrated in Figure~\ref{fig:general-schema}, two web data sets are interlinked using a method for comparing their resources. 
We do not specify at this stage if the method should be automatic or manual. 
Neither do we specify if the two data sets are described using a common ontology or if the ontologies describing their resources differ. 

This is the goal of the following subsections to consider this.
We first consider each case that may happen when interlinking data and describe them abstractly and through an example.
In the end, we unify all this cases in a common framework. 

\subsection{Manual interlinking}\label{sec:manual}

\begin{figure}[!ht]
 \centering
 \includegraphics{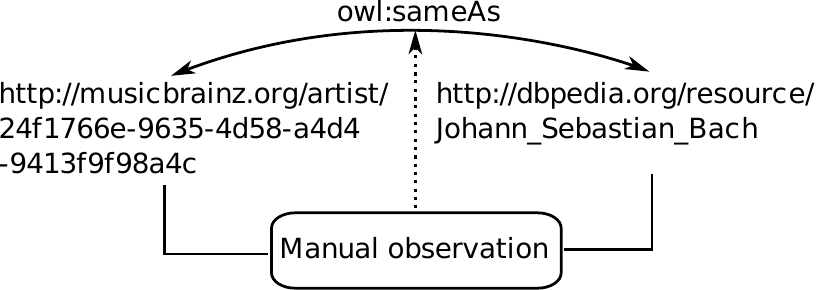}
 \caption{Example of manually linked resources.}
 \label{fig:manual-matching-example}
\end{figure}

In the first case, illustrated in Figure~\ref{fig:manual-matching-example}, resources are manually interlinked.

Manually linking resources can be performed using collaborative tools in the case of large data sets.

\subsection{URI correspondence.}\label{sec:uricorresp}

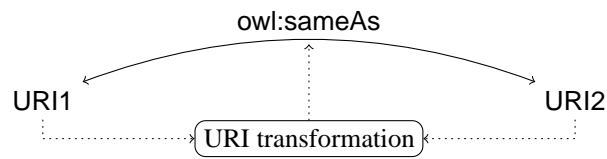
\begin{figure}[!ht]
\begin{center}
\begin{tikzpicture}[cap=round]

\draw (0,3) node (u1) {\textsf{URI1}};
\draw (7,3) node (u2) {\textsf{URI2}};

\draw (3.5,2.5) node[draw,rounded corners] (di) {URI transformation};

\draw[<->] (u1.north east) .. controls +(2cm,7.5mm) and +(-2cm,7.5mm) .. node[above] {\textsf{owl:sameAs}} (u2.north west);
\draw[->,dotted] (u1) |- (di);
\draw[->,dotted] (u2) |- (di);
\draw[->,dotted] (di) -- (3.5,3.75);

\end{tikzpicture} 
\end{center}
 \caption{Data interlinking through URI transformation.}
 \label{fig:uri-alignment}
\end{figure}

In some cases, illustrated in Figure~\ref{fig:uri-alignment}, resources can be trivially linked using a simple transformation of their URIs. 

\begin{figure}[!ht]
 \centering
 \includegraphics{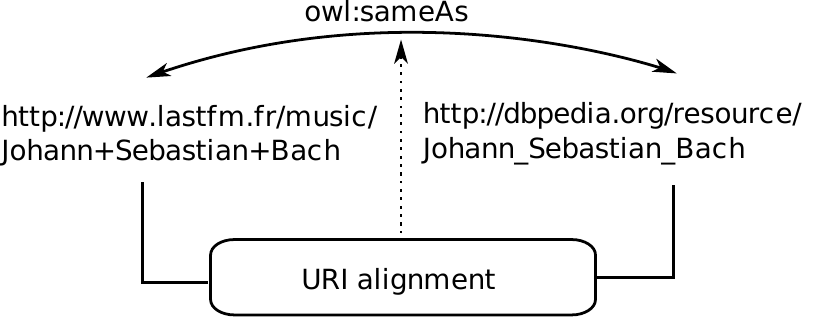}
 \caption{Example of resource linking using the correspondence between URIs.}
 \label{fig:uri-alignment-example}
\end{figure}

A set of rules can be defined to identify equivalent resources from their identifier. For example, in the data set LastFM\footnote{\url{http://last.fm}}, the URI representing an artist is built on the pattern ``First\_name+Last\_name''. Person URIs in DBpedia\footnote{\url{http://dbpedia.org}} are built around the pattern ``FirstName\-\_LastName''. A trivial algorithm can be developed to find equivalent artists based on their URIs. This is illustrated in Figure~\ref{fig:uri-alignment-example} for the composer J.S. Bach.

\subsection{Datasets sharing the same ontologies}\label{sec:sharedonto}

\begin{figure}[!ht]
\begin{center}
\begin{tikzpicture}[cap=round]

\draw (3.5,1) node (o1) {\textsf{O1}};
\draw (0,3) node (u1) {\textsf{URI1}};
\draw (7,3) node (u2) {\textsf{URI2}};

\draw (3.5,2.5) node[draw,rounded corners,text width=3cm, text badly centered] (di) {Resource matching of data sets described by the same ontology};

\draw[->] (o1.west) .. controls +(-2cm,0.5cm) and +(-0.5cm,-0.5cm) ..  (u1);
\draw[->] (o1.east) .. controls +(2cm,0.5cm) and +(0.5cm,-0.5cm) ..  (u2);

\draw[<->] (u1.north east) .. controls +(2cm,7.5mm) and +(-2cm,7.5mm) .. node[above] {\textsf{owl:sameAs}} (u2.north west);
\draw[->,dotted] (u1) |- (di.west);
\draw[->,dotted] (u2) |- (di.east);
\draw[->,dotted] (o1) |- (di.south);
\draw[->,dotted] (di) -- (3.5,3.75);

\end{tikzpicture} 
\end{center}
 \caption{Interlinking two data sets described according to the same ontology.}
 \label{fig:one-ontology}
\end{figure}
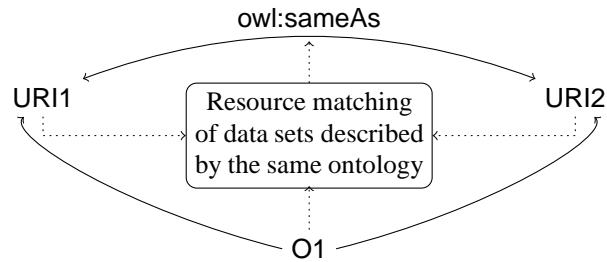

Further than URIs, it may be necessary to consider the ontologies in order to identify entities.
In a first case, 
illustrated in Figure~\ref{fig:one-ontology}, 
the two data sets to interlink are described by the same ontology. The role of the interlinking system is to analyze resources of the same type in order to detect the equivalent ones. To do this, the system compares resource properties with a similarity measure. Systems in this category take as input the properties to compare, the type of comparison algorithm to use for each property, and the method to aggregate the similarity measures of the various properties in order to construct a measure between two resources.

\begin{figure}[!ht]
\begin{center}
 \includegraphics{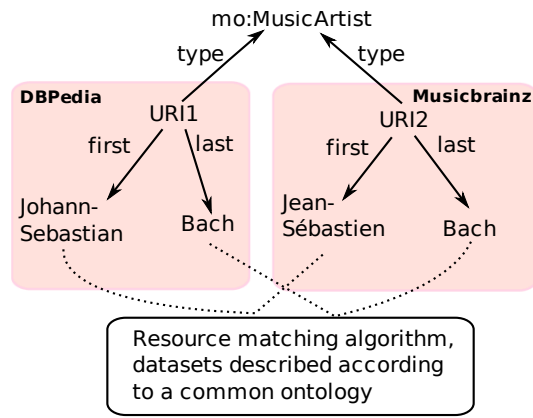}
\end{center}
 \caption{Example of linking resources described according to the same ontology.}
 \label{fig:one-ontology-example}
\end{figure}

For example, Jamendo\footnote{\url{http://www.jamendo.com}} and MusicBrainz\footnote{\url{http://www.musicbrainz.org}}, two data sets containing musicological data, are both described according to a common music ontology \cite{raimond:2007}. The artist J.S. Bach can be identified in both data sets by observing the first name and last name properties of the class \textit{MusicArtist}. It is not possible in this case to identify the equivalence of resources based on their URIs. This example is illustrated in Figure~\ref{fig:one-ontology-example}.

\subsection{Datasets described with heterogeneous ontologies}\label{sec:implicitalign}

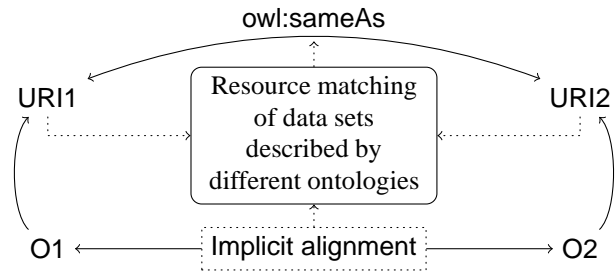
\begin{figure}[!ht]
\begin{center}
\begin{tikzpicture}[cap=round]

\draw (0,1) node (o1) {\textsf{O1}};
\draw (7,1) node (o2) {\textsf{O2}};
\draw (0,3) node (u1) {\textsf{URI1}};
\draw (7,3) node (u2) {\textsf{URI2}};

\draw (3.5,1) node[draw,dotted] (al) {\textsf{Implicit alignment}};
\draw (3.5,2.5) node[draw, rounded corners,text width=3cm, text badly centered] (di) {Resource matching of data sets described by different ontologies};

\draw[->] (o1) .. controls +(-0.5cm,0.5cm) and +(-0.5cm,-0.5cm) ..  (u1);
\draw[->] (o2) .. controls +(0.5cm,0.5cm) and +(0.5cm,-0.5cm) ..  (u2);

\draw[<->] (u1.north east) .. controls +(2cm,7.5mm) and +(-2cm,7.5mm) .. node[above] {\textsf{owl:sameAs}} (u2.north west);
\draw[->,dotted] (u1) |- (di.west);
\draw[->,dotted] (u2) |- (di.east);
\draw[->,dotted] (al) -- (di.south);
\draw[->,dotted] (di) -- (3.5,3.75);
\draw[->] (al) -- (o1);
\draw[->] (al) -- (o2);

\end{tikzpicture} 
\end{center}
 \caption{Two data sets interlinked using an implicit alignment.}
 \label{fig:two-ontologies-implicit}
\end{figure}

Datasets can be described by different ontologies.
This case is illustrated in Figure~\ref{fig:two-ontologies-implicit}.
In order to know which types of entities have to be linked together, the system needs to know the correspondences between these types of entities. 
Then it can work similarly as if there were a single ontology. 

We represent this case in Figure~\ref{fig:two-ontologies-implicit} by introducing the correspondences between ontology classes as an alignment. 
This alignment is presented as implicit because it does not exist as such, but it is mixed with the linking specification or the data interlinking system.


Consider two data sets, one described using FOAF, the other using VCard. 
The linking specification will indicate to the tool to compare entities of type \smtt{foaf:Person} and entities of type \smtt{vcard:VC}, and that when comparing resources of these types, the properties \smtt{foaf:givenname} should be compared to \smtt{vcard:fn}, as well as the property \smtt{foaf:familyname} compared to the property \smtt{vcard:ln}.
This is an implicit alignment containing two correspondences.

\begin{figure}[!ht]
 \centering
 \includegraphics{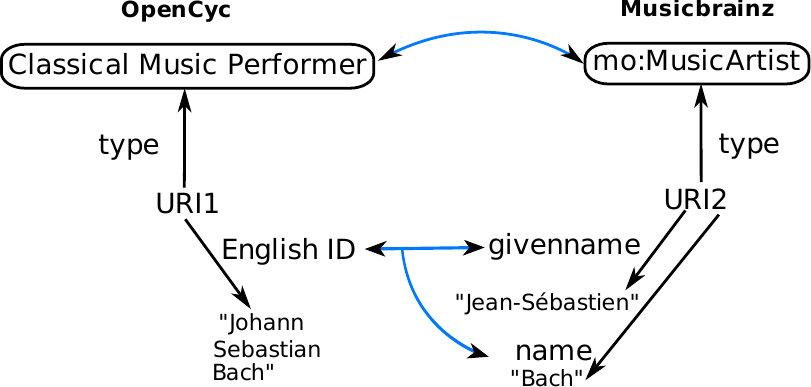}
 \caption{Example of two data sets described with heterogeneous ontologies.}
 \label{fig:two-ontologies-example}
\end{figure}

For example, OpenCyc\footnote{\url{sw.opencyc.org}} represents the artist J.S. Bach using a different ontology than the one used to describe MusicBrainz. The properties ``firstname'' and ``lastname'' correspond to a property ``EnglishID'' in which both names are concatenated. The class \textit{MusicArtist} in the Music Ontology corresponds to a class \textit{Classical Music Composer} in OpenCyc. An alignment between classes and properties needs to be specified in order to find an equivalence between the two resources. This example is illustrated in Figure~\ref{fig:two-ontologies-example}.

\subsection{Data interlinking with alignments}\label{sec:explicitalign}

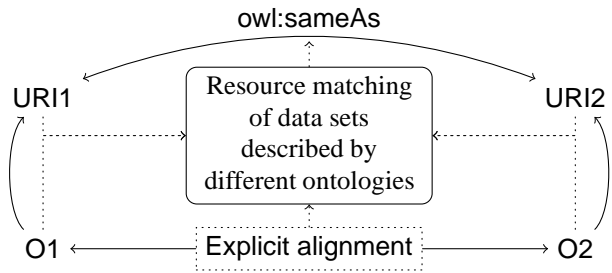
\begin{figure}[!ht]
\begin{center}
\begin{tikzpicture}[cap=round]

\draw (0,1) node (o1) {\textsf{O1}};
\draw (7,1) node (o2) {\textsf{O2}};
\draw (0,3) node (u1) {\textsf{URI1}};
\draw (7,3) node (u2) {\textsf{URI2}};

\draw (3.5,1) node[draw,dotted] (al) {\textsf{Explicit alignment}};
\draw (3.5,2.5) node[draw,rounded corners,text width=3cm, text badly centered] (di) {Resource matching of data sets described by different ontologies};

\draw[->] (o1) .. controls +(-0.5cm,0.5cm) and +(-0.5cm,-0.5cm) ..  (u1);
\draw[->] (o2) .. controls +(0.5cm,0.5cm) and +(0.5cm,-0.5cm) ..  (u2);

\draw[<->] (u1.north east) .. controls +(2cm,7.5mm) and +(-2cm,7.5mm) .. node[above] {\textsf{owl:sameAs}} (u2.north west);
\draw[->,dotted] (u1) |- (di);
\draw[->,dotted] (u2) |- (di);
\draw[->,dotted] (o1) |- (di);
\draw[->,dotted] (o2) |- (di);
\draw[->,dotted] (al) -- (di);
\draw[->,dotted] (di) -- (3.5,3.75);
\draw[->] (al) -- (o1);
\draw[->] (al) -- (o2);

\end{tikzpicture} 
\end{center}
 \caption{Two data sets matched using an explicit alignment.}
 \label{fig:two-ontologies-explicit}
\end{figure}

Another approach, illustrated in Figure~\ref{fig:two-ontologies-explicit}, takes advantage of an already existing  explicit alignment between the two ontologies used by the data sets.

An additional possibility, not found in existing systems, would be for the data linking system to first match the two ontologies before using the resulting alignment for supporting data interlinking.
In such a system, ontology matching and data interlinking would be merged.

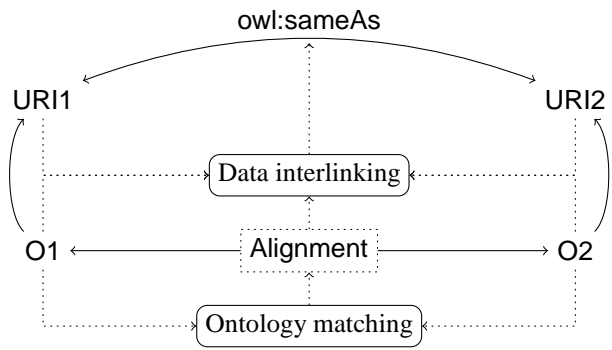
\begin{figure}[!ht]
\begin{center}
\begin{tikzpicture}[cap=round]

\draw (0,1) node (o1) {\textsf{O1}};
\draw (7,1) node (o2) {\textsf{O2}};
\draw (0,3) node (u1) {\textsf{URI1}};
\draw (7,3) node (u2) {\textsf{URI2}};

\draw (3.5,0) node[draw,rounded corners] (om) {Ontology matching};
\draw (3.5,1) node[draw,dotted] (al) {\textsf{Alignment}};
\draw (3.5,2) node[draw,rounded corners] (di) {Data interlinking};

\draw[->] (o1) .. controls +(-0.5cm,0.5cm) and +(-0.5cm,-0.5cm) ..  (u1);
\draw[->] (o2) .. controls +(0.5cm,0.5cm) and +(0.5cm,-0.5cm) ..  (u2);

\draw[<->] (u1.north east) .. controls +(2cm,7.5mm) and +(-2cm,7.5mm) .. node[above] {\textsf{owl:sameAs}} (u2.north west);
\draw[->,dotted] (o1) |- (om);
\draw[->,dotted] (o2) |- (om);
\draw[->,dotted] (om) -- (al);
\draw[->,dotted] (u1) |- (di);
\draw[->,dotted] (u2) |- (di);
\draw[->,dotted] (o1) |- (di);
\draw[->,dotted] (o2) |- (di);
\draw[->,dotted] (al) -- (di);
\draw[->,dotted] (di) -- (3.5,3.75);
\draw[->] (al) -- (o1);
\draw[->] (al) -- (o2);

\end{tikzpicture} 
\end{center}
\caption{General framework for data interlinking involving ontology matching.}\label{fig:framework}
\end{figure}

\bigskip
Figure~\ref{fig:framework} unifies all these processes in a single description.
This framework leads to clarify interactions between data interlinking and ontology matching. 

The next section discusses different systems and their position with respect to the proposed framework.

\section{Data interlinking tool analysis}\label{sec:analysis}

The work presented in this section is the result of the MeLinDa experiment conducted jointly 
with the linked open data mailing list.
We asked interlinking tool developers to send us the linking specifications their tools take as input. We then compared these specifications and evaluated the possibility to publish them in a common language\footnote{\url{http://melinda.inrialpes.fr}. MeLinDa stands for Meta-Linking Data.}. Six systems took part in the experiment. We are aware of at least two other systems not analyzed in this study \cite{sais:2008,hogan:2007}.

We present below different criteria along which the tools can be compared, then we briefly describe the specifics of each tool and provide comparison of them along the criteria (Table~\ref{tab:datalinkers}).

\subsection{Analysis criteria}\label{sec:criteria}

For each analyzed tool, we tried to answer several questions reproduced below. 
We will then describe and categorize each tool according to these questions.
\begin{description}
 \item[Degree of automation]~
 \begin{seconditemize}
  \item Is the tool completely automatic (a black box)?
  \item Does the tool need to be parametrized by the user? What kind of parameters (data matching techniques, ontology alignment)?
 \end{seconditemize}
\item[Used matching techniques]~
 \begin{seconditemize}
 \item String matching?
 \item External functions (values conversion, data transformations)?
 \item Similarity propagation?
 \item Other techniques?
 \end{seconditemize}
\item[Access]~
\begin{seconditemize}
 \item How does the tool access data? (SPARQL endpoint, RDF Dump, URL)
\end{seconditemize}
\item[Ontologies]~
\begin{seconditemize}
 \item Does the tool take into account ontologies associated to the data sets?
 \item Does the tool allow to interlink data sets described according to different ontologies?
 \item If the ontologies differ, does the tool perform ontology matching?
\end{seconditemize}
\item[Output]~
\begin{seconditemize}
 \item What does the tool produce as output (\smtt{sameAs} links, VoiD linkset, other type of links)?
 \item Does the tool propose to merge the two input data sets?
\end{seconditemize}
\item[Domain] Is the tool specific for a given domain?
\item[Post-processing] Does the tool perform any post-processing operations (consistency checking and inconsistency resolution)?
\end{description}

\subsection{Tools}

We considered the 6 following tools. Table~\ref{tab:datalinkers} summarizes the analysis.

\begin{sidewaystable}[!htbp]
\begin{center}
 \begin{tabular}{|l|l|l|l|l|l|l|}
 \hline
                    & \textbf{RKB CRS} &\textbf{LD-Mapper}  & \textbf{ODD} & \textbf{RDF-AI} & \textbf{Silk}  & \textbf{Knofuss} \\ \hline
 Ontologies         & multiple & multiple& single             & single & single         & multiple               \\
                    & & & & & &   \\ \hline
 Automation     & semi- & automatic   & semi- & semi- & semi- & semi- \\               
 &automatic&&automatic&automatic&automatic&automatic\\ \hline
 User input  & program  & none  & link spec. & data set structure & links spec. & fusion onto.  \\
             &  &     & query & alignment method & alignment method &  \\ \hline
 Input format   & Java       & prolog         & LinQL & XML             & Silk-LSL (XML) & OWL \\ 
 &&&&&&\\ \hline
 Matching & string & string,  & string & string, & string & string, \\
 techniques     & & similarity& & Wordnet & & adaptive \\
                   &          &  propagation  & &          & & learning \\ \hline
 Onto. alignment & no  & no  & no & no   & no & yes, in input\\         
 &&&&&& \\ \hline
 Output     & owl:sameAs      & owl:sameAs     & linkset & alignment format  & alignment format & alignment format \\
             & &   linkset &                &merged data set& linkset & merged data set\\ 
             &        &                    && &&\\  \hline
 Data access  & API             & local copy & ODBC & local copy & SPARQL & local copy   \\ 
 &&&&& & \\ \hline
 Domain  & publications   & Music Ontology  & independent & independent & independent & independent  \\ 
 &&&&&& \\ \hline
 Post-processing & no  & no & no & no & no & inconsistency  \\
                    &                     & & & & &  resolution \\\hline
 \end{tabular}
 \end{center}
\caption{Comparison of data linking tools.}\label{tab:datalinkers}
 \end{sidewaystable}
 
\subsubsection{RKB-CRS}
The co-reference resolution system (CRS) of the RKB knowledge base \cite{jaffri:2008b} is built around URI equivalence lists. These lists are built using a Java program working on the specific domain of universities and conferences. A new program needs to be written for each pair of data sets to integrate. Each program consists of the selection of the resources to compare, and their comparison using string similarity on the resource property values.

\subsubsection{LD-Mapper}
LD-Mapper \cite{raimond:2008} is a data set integration tool working on the music domain. This tool is based on a similarity aggregation algorithm taking into account the similarity of a resource's neighbors in the graph describing it. It requires little user configuration but only works with data sets described with the Music Ontology \cite{raimond:2007}. LD-Mapper is implemented in Prolog.

\subsubsection{ODD-linker}
ODD-Linker \cite{hassanzadeh:2009} is an interlinking tool implemented on top of a tool mining for equivalent records in relational databases. ODD-Linker uses SQL queries for identifying and comparing resources. The tool translates link specifications expressed in the LinQL dedicated language originally developed for duplicate records detection in relational databases. Its usage in the context of linked data is thus limited to relational databases exposed as linked data. LinQL is nonetheless an expressive formalism for link specifications. The language supports many string matching algorithms, hyponyms and synonyms, conditions on attribute values. 

\subsubsection{RDF-AI}
RDF-AI \cite{scharffe:2009} is an architecture for data set matching and fusion. It generates an alignment that can be later used either to generate a linkset, or to merge two data sets. The interlinking parameters of RDF-AI are given in a set of XML files corresponding to the different steps of the process. The data set structure and the resources to match are described in two files. This description corresponds to a small ontology containing only resources of interest and the properties to use in the matching process. A pre-processing file describes operations to perform on resources before matching. Translation of properties and name reordering are performed before looking for links. A matching configuration file describes which techniques should be used for which resources. A threshold for generating the linkset from the alignment can be specified. Additionally, when data sets need to be merged, a configuration file describes the fusion method to use. The prototype works with a local copy of the data sets and is implemented in Java.
 
\subsubsection{Silk}
Silk \cite{bizer:2009} is an interlinking tool parametrized by a link specification language: the Silk Link Specification Language (Silk LSL, see \S\ref{sec:silk}). The user specifies the type of resources to link and the comparison techniques to use. Datasets are referenced by giving the URI of the SPARQL endpoint from which they are accessible. A named graph can be specified in order to link only resources belonging to this graph. Resources to be linked are specified using their type, or the RDF path to access them. Silk uses many string comparison techniques, numerical and date similarity measures, concept distances in a taxonomy, and set similarities.  A condition allows for specifying the matching algorithm used to match resources. Matching algorithms can be combined using a set of operators (MAX, MIN, AVG) and literals can be transformed before the comparison by specifying a transformation function, concatenating or splitting resources. Regular expressions can be be used to preprocess resources. Silk takes as input two web data sets accessible behind a SPARQL endpoint. When resources are matched with a confidence over a given threshold, the tool outputs \smtt{sameAs} links or any other RDF predicate specified by the user. The first version of Silk was implemented in Python; version~2 is a new implementation in Scala.

\subsubsection{Knofuss}
The Knossos architecture \cite{nikolov:2008} aims at providing support for data set fusion. A specificity of Knofuss is the possibility to use existing ontology alignments. The resource comparison process is driven by a dedicated ontology for each data set specifying resources to compare, as well as the comparison techniques to use. 
Each ontology gives, 
for each type of resource to be matched, an \textit{application context} defined as a SPARQL query for this type of resource. An \textit{object context model} is also defined to specify properties that will be used to match these resource types. Corresponding application contexts are given the same ID in the two ontologies and one application context indicates which similarity metric should be used for comparing them. When the two data sets are described using different ontologies, an ontology alignment can be specified. This alignment is given in the ontology alignment format \cite{david2011a}. Knofuss allows for exporting links between data sets, but was originally designed to merge equivalent resources. It includes a consistency resolution module which ensures that the data sets resulting from the fusion of the two data sets is consistent with respect to the ontologies. The parameters of the fusion operation are also given in the input ontologies. Knofuss works with local copies of the data sets and is implemented in Java.

\bigskip
An analysis of these tools according to the criteria of Section~\ref{sec:criteria} is summarized in Table~\ref{tab:datalinkers}.
Obviously there is a lot of variation between these tools in spite of their common goal.
Even if they are very diverse,  each of these data interlinking tools fit in the proposed framework as shown on Table~\ref{tab:systems}.

\begin{table}[!h]
\begin{center}
\begin{tabular}{|l|l|}
\hline
Case & Tool\\ \hline
Manual link specification (\S\ref{sec:manual}) & \\ \hline
URI correspondence (\S\ref{sec:uricorresp}) & RKB-CRS \\ \hline
Common ontology (\S\ref{sec:sharedonto}) & LD-Mapper, ODD-Linker\\ \hline
Different ontologies, 
implicit alignment (\S\ref{sec:implicitalign}) & RDF-AI, Silk \\ \hline
Different ontologies, 
explicit alignement (\S\ref{sec:explicitalign}) & Knofuss \\ \hline
\end{tabular}
\caption{Classification of analyzed tools with regard to the framework.}\label{tab:systems}
\end{center}
\end{table}

The goal of the next section is to consider 
how using ontology alignments could lead to more automation for the interlinking task, as well as how linked data could provide evidence for obtaining better ontology alignments.

\section{Matching/linking cooperation}\label{sec:coop}
 

Although ontology matching and data interlinking can be similar at a certain level (they both relate formal entities), there are important differences: one acts at the schema level and the other at the instance level. 
In fact, ontology matching can take advantage of linked data as an external source of information for ontology matching, and, conversely, data interlinking can benefit from ontology matching by using correspondences to focus the search for potential instance level links.

These differences are reflected in the types of specification involved in these processes:
\begin{itemize*}
\item A link, e.g., a \smtt{sameAs} statement, tells which \smtt{City} in wikipedia correspond to which \smtt{P} (place) in geonames, e.g., Manchester sameAs Manchester.
\item A linking specification tells how to find the former, e.g., for linking a \smtt{City} to a \smtt{P}, evaluate how the \smtt{label} of the first one is close to the \smtt{name} of the second one with some measure, e.g., \smtt{jaroSimilarity}, evaluate how the \smtt{populationTotal} of the first one is close to the \smtt{population} of the second one with another measure, e.g., \smtt{numSimilarity}, average the two values and if the result is above .9, then generate the \smtt{sameAs} statement.
\item An ontology alignment tells which components from one ontology corresponds to which components in the other. For example, \smtt{dbpedia:City} is a kind of \smtt{geonames:P} and in this context, \smtt{label} is equivalent to \smtt{name} and \smtt{populationTotal} is equivalent to \smtt{population}.
\end{itemize*}
This results in two process specifications -- interlinking and matching -- and their results -- linksets between data and alignments between ontologies. The situation is summarized by Table~\ref{tab:process-results}.
\begin{table}[!ht]
\begin{center}
\begin{tabular}{|c|c|c|}
\hline
& process & result \\
\hline
instance & linking specification & linkset\\
\hline
class & matcher & alignment \\
\hline
\end{tabular}
\end{center}
\caption{Interlinking and matching processes and their results.}
\label{tab:process-results}
\end{table}

By clearly establishing these differences, we obtain a natural partitioning between data links, linking specifications and ontology alignments and the languages for expressing them:
\begin{description}
 \item[The assertion expression language] (e.g., RDF and VoiD) allows for representing equivalence between resources in data sets;
 \item[The linking specification language] (e.g., Silk, LinQL) allows for defining how to search for equivalence between resources;
 \item[The alignment representation language] (e.g., the Alignment format or EDOAL) allows for specifying equivalence rules between ontological entities.
\end{description}

It would be useful to take advantage of the framework of Section~\ref{sec:framework}, to help tools interoperate. 
This would present many advantages, in particular the possibility to share, distribute and improve linking specifications, as well as reuse them or extend them instead of computing them again whenever a data set is modified. 
This would also allow to compose linking specifications such that it would be possible to go from one data set to another without going through an intermediary. 

We propose a scheme under which it is possible for data linking tools to take ontology alignments as a way to constrain their solution space.
Figure~\ref{fig:framework} provides a natural way to implement this collaboration.

We first present an expressive language for ontology alignments 
that can be exploited by data interlinking systems (Section~\ref{sec:edoal}).
and briefly introduce the linking specification language Silk-LSL (Section~\ref{sec:silk}).
Then we show how they could fruitfully be combined for data interlinking (Section~\ref{sec:extsilk}).


\section{EDOAL: an expressive ontology alignment language}\label{sec:edoal}

EDOAL (Expressive Declarative Ontology Alignment Language) is the new name of the OMWG mapping language for expressing ontology alignment \cite{euzenat:2007} that has been available through the Alignment API since version 3.1.
This language is an extension of the Alignment format \cite{euzenat:2004} that can be generated by most matchers.
Its main purpose is to offer more expressiveness in the way alignments are expressed. 
It presents the advantage to be declarative and also to specify transformations like those needed in order to construct links between resources. 

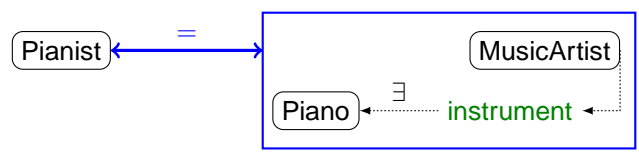
\begin{figure}[ht]
\begin{center}
\begin{tikzpicture}[cap=round]
\ontoset;

\draw (0,2) node[style=class,anchor=west] (t1) {Pianist};


\draw (8,2) node[style=class,anchor=east] (s1) {MusicArtist};
\draw(7.5,1.2) node[style=attr,anchor=east] (size) {instrument}; 

\draw (4,1.2) node[style=class] (sc) {Piano};

\draw [style=const] (s1.east) |- (size.east);
\draw [style=const] (size.west) -- node[above] {$\exists$}  (sc.east);

\draw[color=blue,thick] (3.3,.7) rectangle (8.2,2.5);
\draw (3.3,2) node[anchor=west] (aaa) {};


\draw [style=corresp] (t1.east) -- node[above] {$=$} (aaa.west);

\end{tikzpicture} 
\end{center}
\caption{Correspondence between non named resources.}\label{fig:piano}
\end{figure}

A first advantage of the expressiveness of EDOAL is the possibility to express correspondences between non named entities.
For instance, a simple assertion such as ``a pianist is a musician who plays piano'', can be expressed by (Figure~\ref{fig:piano}):
\begin{smalltt}
:dbp-mo a align:Alignment;
    align:onto1 <http://dbpedia.org/ontology/>;
    align:onto2 <http://www.musicontology.com/>;
    align:map [ :map1 a align:Cell;
      align:entity1 dbp:Pianist;
      align:entity2 [ a edoal:Class;
        edoal:and mo:MusicArtist;
        edoal:and [ a edoal:PropertyValueConstraint;
          edoal:property mo:instrument;
          edoal:value mo:Piano.
        ].
      ];
      align:relation align:equivalent;
    ]. 
\end{smalltt}

This can help restricting the search space of data interlinking tools far beyond what they currently do (named classes).

\begin{figure}[!ht]
\begin{center}
\begin{tikzpicture}[cap=round]
\ontoset;

\draw (0,2.5) node[style=class,anchor=west] (t1) {Classical Music Performer};
\draw(0.3,1.2) node[style=attr,anchor=west] (year) {rdf:id}; 

\draw [style=const] (t1.west) |- (year.west);


\draw (8,2.5) node[style=class,anchor=east] (s1) {MusicArtist};
\draw(7.7,1.2) node[style=attr,anchor=east] (size) {rdf:id}; 

\draw [style=const] (s1.east) |- (size.east);


\draw [style=corresp] (t1.east) -- node[above] (a) {$\leq$} (s1.west);
\draw [style=corresp,dashed] (year.east) -- node[below]{http://.../([\^{}\_]*)\$1\_([\^{}.]*])\$2.rdf} node[above]{http://.../([\^{}+]*)\$2+([\^{}/]*)\$1/} (size.west);
\end{tikzpicture} 
\end{center}
\caption{Expression of a resource equivalence represented in an expressive ontology alignment language.}\label{fig:corr5}
\end{figure}
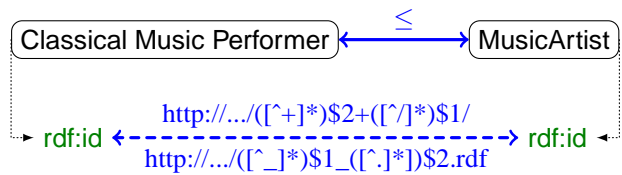

In addition, in EDOAL, it is possible to express that two classes are equivalent, and that their instances are equivalent modulo a transformation.
This can be used for covering, without further information, the URI correspondence case of the framework (Section~\ref{sec:uricorresp}).
For instance, Figure~\ref{fig:corr5} shows an EDOAL correspondence using regular expression transformations for identifying musician instances between two data sets with different conventions.

Of course, this can only work when there exists such correspondences, i.e., an exact method for generating links.
Most of the time, data interlinking systems still need to use heuristics to find links between entities.
This can be provided by the simple Alignment format, but EDOAL can do more by indicating where to look for to establish the correspondence.

\begin{figure}[!ht]
\begin{center}
\begin{tikzpicture}[cap=round]
\ontoset;

\draw (0.2,2) node[style=class,anchor=west] (t1) {City};
\draw(0.7,1.4) node[style=attr,anchor=west] (name1) {rdfs:label}; 
\draw(0.7,0.8) node[style=attr,anchor=west] (pop1) {populationTotal}; 

\draw [style=const] (t1.west) |- (name1.west); 
\draw [style=const] (t1.west) |- (pop1.west); 

\draw[color=blue,thick,dashed] (0,0.5) rectangle (3.5,2.5);


\draw (7.8,2) node[style=class,anchor=east] (s1) {P};
\draw(7.3,1.4) node[style=attr,anchor=east] (name) {name}; 
\draw(7.3,0.8) node[style=attr,anchor=east] (pop) {population}; 

\draw [style=const] (s1.east) |- (name.east); 
\draw [style=const] (s1.east) |- (pop.east); 

\draw[color=blue,thick,dashed] (5,0.5) rectangle (8,2.5);


\draw [style=corresp] (t1.east) -- node[above] {$\leq$} (s1.west);
\draw [style=corresp] (name1.east) -- node[above] {=}  (name.west);
\draw [style=corresp] (pop1.east) -- node[above] {=}  (pop.west);

\end{tikzpicture} 
\end{center}
\caption{Contextual matching of of two classes and its properties.}\label{fig:contextmatch}
\end{figure}
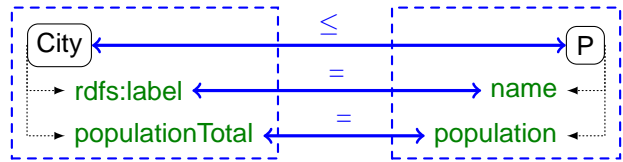

In particular, EDOAL allows for expressing contextual relations between elements.
For instance, the typical example in Silk documentation is the linking of DBpedia cities and geoname P(laces) 
through comparing their names and populations. 
Expressing this with the simple alignment:
\begin{smalltt}
:dbp-geo a align:Alignment;
    align:onto1 <http://dbpedia.org/ontology/>;
    align:onto2 <http://www.geonames.org/ontology#>;
    align:map [ :map1 a align:Cell;
      align:entity1 dbpedia:City;
      align:entity2 gn:P;
      align:relation align:subsumedBy.
    ];
    align:map [ :map2 a align:Cell;
      align:entity1 dbpedia:populationTotal;
      align:entity2 gn:population;
      align:relation align:equivalent.
    ];
    align:map [ :map3 a align:Cell;
      align:entity1 rdfs:label;
      align:entity2 gn:name;
      align:relation align:equivalent.
    ]. 
\end{smalltt}
\noindent does not express the expected meaning because, of course, \smtt{rdfs:label} is not equivalent to \smtt{gn:name}.
One could consider expressing that \smtt{gn:name} is more specific than \smtt{rdfs:label}. 
This is correct but still not precise enough.
The intended meaning is that, in the context of \smtt{dbpdia:City} and \smtt{gn:P}, these two properties are equivalent.
This is what EDOAL can express through the schema of Figure~\ref{fig:contextmatch} corresponding to the following alignment:
\begin{smalltt}
:dbp-geo a align:Alignment;
    align:onto1 <http://dbpedia.org/ontology/>;
    align:onto2 <http://www.geonames.org/ontology#>;
    align:map [ :map1 a align:Cell;
      align:entity1 dbpedia:City;
      align:entity2 gn:P;
      align:relation align:subsumedBy.
    ];
    align:map [ :map2 a align:Cell;
      align:entity1 [ a align:Property; 
        edoal:and dbpedia:populationTotal. 
        edoal:and [ a edoal:PropertyDomainRestriction;
          edoal:domain dbpedia:City. ];
      align:entity2 [ a align:Property; 
        edoal:and gn:population; 
        edoal:and [ a edoal:PropertyDomainRestriction;
          edoal:domain gn:P. ];
      align:relation align:equivalent.
    ];
    align:map [ :map2 a align:Cell;
      align:entity1 [ a align:Property; 
        edoal:and rdfs:label.
        edoal:and [ a edoal:PropertyDomainRestriction;
          edoal:domain dbpedia:City. ];
      align:entity2 [ a align:Property; 
        edoal:and gn:name; 
        edoal:and [ a edoal:PropertyDomainRestriction;
          edoal:domain gn:P. ];
      align:relation align:equivalent.
    ]. 
\end{smalltt}

Even if such an alignment would provide information to data interlinking tools, this is still not sufficient.
Of course, it tells which properties should be equivalent and thus can be used for identifying entities.
But it does not tell how to take them into account.
So, this alignement would be sufficient to link entities if the values of \smtt{rdfs:label} were exactly the same as those of \smtt{gn:name} and the values of \smtt{populationTotal} were exactly the same as those of \smtt{population}, but not otherwise.

EDOAL provides more features for transforming this information, like we have seen in Figure~\ref{fig:corr5}.
This could be helpful but the problem is deeper: data interlinking is a decision problem rather that just a transformation.
It is the role of the data linking specification to tell when a particular \smtt{dbpedia:City} and a \smtt{gn:P} should be considered the same. This is why we propose to use data interlinking specifications together with alignments.

\section{Silk-LSL: a linking specification language}\label{sec:silk}

Below is the Silk-LSL \cite{bizer:2009} specification to interlink cities in the two data sets DBpedia and Geonames:

\begin{smalltt}
<Silk>

  <Prefix id="rdfs" namespace=
          "http://www.w3.org/2000/01/rdf-schema#" />
  <Prefix id="dbpedia" namespace=
          "http://dbpedia.org/ontology/" />
  <Prefix id="gn" namespace=
          "http://www.geonames.org/ontology#" />

  <DataSource id="dbpedia">
    <EndpointURI>http://demo_sparql_server1/sparql
    </EndpointURI>
    <Graph>http://dbpedia.org</Graph>
  </DataSource>

  <DataSource id="geonames">
    <EndpointURI>http://demo_sparql_server2/sparql
    </EndpointURI>
    <Graph>http://sws.geonames.org/</Graph>
  </DataSource>

  <Interlink id="cities">
    <LinkType>owl:sameAs</LinkType>

    <SourceDataset dataSource="dbpedia" var="a">
      <RestrictTo>
         ?a rdf:type dbpedia:City
      </RestrictTo>
    </SourceDataset>

    <TargetDataset dataSource="geonames" var="b">
      <RestrictTo>
        ?b rdf:type gn:P
      </RestrictTo>
    </TargetDataset>

    <LinkCondition>
      <AVG>
        <Compare metric="jaroSimilarity">
          <Param name="str1" path="?a/rdfs:label" />
          <Param name="str2" path="?b/gn:name" />
        </Compare>
        <Compare metric="numSimilarity">
          <Param name="num1" 
                 path="?a/dbpedia:populationTotal" />
          <Param name="num2" path="?b/gn:population" />
        </Compare>
      </AVG>
    </LinkCondition>

    <Thresholds accept="0.9" verify="0.7" />
    <Output acceptedLinks="accepted_links.n3"
      verifyLinks="verify_links.n3"
      mode="truncate" />
  </Interlink>

</Silk>
\end{smalltt}

This specification fulfills two roles:
\begin{itemize}
\item It is an alignment: it specifies the classes in which entities to link can be found.
Restrictions to \smtt{dbpedia:City} and \smtt{gn:P} are in fact an alignment between these two concepts. 
Similarly, the compared properties \smtt{populationTotal} and \smtt{population} and  \smtt{rdfs:label} and \smtt{name}, respectively provide the correspondences between properties.
\item It specifies how to link entities. Indeed, what Silk brings in addition is the specification of how to decide if two entities should be linked: when the average (\smtt{AVG}) of their respective distances  (\smtt{Compare}) is over a threshold (\smtt{Threshold}, there are two thresholds, one for accepting automatically the equivalence and one for drawing the attention of a user).
\end{itemize}

It could be possible to refer to an external alignment between the two underlying ontologies instead of specifying it in the linking specification. 
This approach would present obvious reuse advantages when other data sets requiring the same alignment, i.e., using the same ontologies, need to be interlinked.

\section{Data interlinking using ontology alignments}\label{sec:extsilk}

Apart from Knofuss, interlinking tools do not provide the possibility to use an ontology alignment. 
Knofuss still needs to specify queries on both data sets from which results equivalent resources will be identified. 

Indeed, using an explicit alignment, provided that it is expressive enough, can serve two functions:
\begin{enumerate*}
\item narrowing the search space through pointing to equivalent concepts, and
\item providing the properties that can be used for identifying concepts.
\end{enumerate*}

There are two ways to articulate ontology alignment and linking specifications:
\begin{itemize*}
\item Transforming an expressive alignment into a linking specification: this requires that the alignment contains as much information as possible and that matchers be able to produce such descriptions. This has the advantage that from the alignment, the specification may be transformed into different linking specification languages.
\item Enabling linking specifications to refer explicitly to alignments and eventually to matchers: this requires extending specification languages for that purpose.
\end{itemize*}
We consider the latter option below.
Given that the alignment is available, it is possible to simplify the Silk specification and refer to the alignment, by introducing three types of information:
which alignments to use (\smtt{UseAlignment}), entities of which correspondences must be linked (\smtt{LinkCell}) and which matched properties can be compared for identifying entities (\smtt{CellParam}).

\begin{smalltt}

  <UseAlignment rdf:resource="#dbp-geo" />
  
  <Interlink id="cities">
    <LinkType>owl:sameAs</LinkType>
    
    <LinkCell rdf:resource="#map1" />

    <LinkCondition>
      <And combiner="AVG">
        <Compare metric="jaroSimilarity">
          <CellParam rdf:resource="#map2" />
        </Compare>
        <Compare metric="numSimilarity">
          <CellParam rdf:resource="#map3" />
        </Compare>
      </And>
    </LinkCondition>

    <Thresholds accept="0.9" verify="0.7" />
    <Output acceptedLinks="accepted_links.n3"
      verifyLinks="verify_links.n3"
      mode="truncate" />
  </Interlink>
\end{smalltt}

The specifics of the data interlinking task remain in this specification: how to compare values, how to aggregate their results and when to issue the link or not.

In fact, the symbiosis between the alignment and the linking specification can be rendered even more automatic, e.g., by defining default rules for comparing values of a given type, default rules for aggregating metrics, and default threshold rules. 
However, it is also useful that the linking specification designer can keep control on what the interlinking tool does and, even if a correspondence is not in an alignment, be able to define it.

This approach presents several advantages:
\begin{enumerate*}
 \item The link specification is simplified, reducing the manual input;
 \item There is a clear separation between links, linking specification, and ontology alignments;
\item The same alignment can be reused for linking any two data sets described according to the same ontologies.
\end{enumerate*}

\section{Conclusion}\label{sec:conclusion}

Interlinking data sets becomes an even more important problem as their
number quickly increase. In order to scale, the interlinking task has to
be as automated as possible. 

We have studied various existing data interlinking tools and observed the following:
\begin{itemize}
 \item Beyond the variations between these systems, it is possible to define a general framework covering the different levels of expressiveness (ranging from a Prolog program to compositions of linking specifications).
 \item Although there is a relevant similarity with ontology alignment, an ontology alignment language is not enough to express linking specifications, particularly because 
it is not its primarily goal to identify individual entities.
\end{itemize}
We have thus proposed an architecture based on three different languages having each its own precise purpose: expressing links, expressing linking specifications, and expressing ontology alignments.

This architecture can be used in order to organize a better collaboration between ontology matchers and data interlinking tools. This can be achieved with only minimal extensions to existing languages.

In particular, we have illustrated the ontology alignment part with EDOAL, an expressive ontology alignment language that offers the necessary concepts for being used in data interlinking. 
On the data interlinking side, we have focussed on the Silk-LSL language which seems to be at once declarative and powerful enough to express a wide range of constraints on data interlinking.
Extending it with the capacity to benefit from ontology alignments would allow tools using it to benefit from the wide range of ontology alignment techniques and tools.

The domain of interlinking data on the web is quickly expanding. 
New needs and new techniques appear. 
It is thus important not to breed innovations with a narrow language. 
Developing standard tools to share link specifications will greatly improve those techniques. 
There is still a lot of work to do in order to achieve this goal.



\section*{Acknowledgements}

We thank the data interlinking tool developers who provided us with linking specifications for their tools.

This work was conducted in the context of the Datalift project funded by the French ANR (Agence Nationale de la Recherche) under grant ANR-10-CORD-009.

\bibliographystyle{named}
\bibliography{bibliography}

\begin{thebibliography}{}

\bibitem[\protect\citeauthoryear{Adida \bgroup \em et al.\egroup
  }{2008}]{adida2008b}
Ben Adida, Mark Birbeck, Shane McCarron, and Steven Pemberton.
\newblock {RDFa} in {XHTML}: Syntax and processing.
\newblock Recommendation, W3C, 2008.
\newblock http://www.w3.org/TR/rdfa-syntax/.

\bibitem[\protect\citeauthoryear{Alexander \bgroup \em et al.\egroup
  }{2009}]{alexander:2009}
Keith Alexander, Richard Cyganiak, Michael Hausenblas, and Jun Zhao.
\newblock Describing linked datasets - on the design and usage of void, the
  'vocabulary of interlinked datasets'.
\newblock In {\em Linked Data on the Web Workshop (LDOW09), Workshop at 18th
  International World Wide Web Conference (WWW09)}, Madrid, Spain, 2009.

\bibitem[\protect\citeauthoryear{Berners-Lee}{2009}]{bernerslee:2009}
Tim Berners-Lee.
\newblock Linked-data design issues.
\newblock W3C design issue document, June 2009.
\newblock http://www.w3.org/DesignIssues/LinkedData.html.

\bibitem[\protect\citeauthoryear{Bizer \bgroup \em et al.\egroup
  }{2009}]{bizer:2009}
Christian Bizer, Julius Volz, Georgi Kobilarov, and Martin Gaedke.
\newblock Silk - a link discovery framework for the web of data.
\newblock In {\em 18th International World Wide Web Conference}, April 2009.

\bibitem[\protect\citeauthoryear{Bizer}{2003}]{bizer:2003}
Christian Bizer.
\newblock D2r map - a database to rdf mapping language.
\newblock In {\em Proc. 12th WWW conference poster session}, 2003.

\bibitem[\protect\citeauthoryear{Bouquet \bgroup \em et al.\egroup
  }{2008}]{bouquet:2008}
Paolo Bouquet, Heiko Stoermer, and Barbara Bazzanella.
\newblock {An Entity Naming System for the Semantic Web}.
\newblock In {\em Proceedings of the 5th European Semantic Web Conference
  (ESWC2008)}, LNCS, June 2008.

\bibitem[\protect\citeauthoryear{{David} \bgroup \em et al.\egroup
  }{2011}]{david2011a}
{Jérôme} {David}, {Jérôme} {Euzenat}, {François} {Scharffe}, and {Cássia}
  {Trojahn dos Santos}.
\newblock The {Alignment API} 4.0.
\newblock {\em Semantic web journal}, 2(1):3--10, 2011.

\bibitem[\protect\citeauthoryear{Elmagarmid \bgroup \em et al.\egroup
  }{2007}]{elmagarmid:2007}
Ahmed Elmagarmid, Panagiotis Ipeirotis, and Vassilios Verykios.
\newblock Duplicate record detection: A survey.
\newblock {\em IEEE Transactions on Knowledge and Data Engineering},
  19(1):1--16, January 2007.

\bibitem[\protect\citeauthoryear{Euzenat and Shvaiko}{2007}]{euzenat:2007b}
J{\'e}r{\^o}me Euzenat and Pavel Shvaiko.
\newblock {\em Ontology matching}.
\newblock Springer-Verlag, Heidelberg (DE), 2007.

\bibitem[\protect\citeauthoryear{Euzenat \bgroup \em et al.\egroup
  }{2007a}]{euzenat:2007c}
J{\'e}r{\^o}me Euzenat, Adrian Mocan, and Fran\c{c}ois Scharffe.
\newblock {\em Ontology Management: Semantic Web, Semantic Web Services, and
  Business Applications}, chapter Ontology Alignments:an ontology management
  perspective.
\newblock Springer, 2007.

\bibitem[\protect\citeauthoryear{Euzenat \bgroup \em et al.\egroup
  }{2007b}]{euzenat:2007}
J{\'e}r{\^o}me Euzenat, Fran\c{c}ois Scharffe, and Antoine Zimmermann.
\newblock D2.2.10: Expressive alignment language and implementation.
\newblock Project deliverable 2.2.10, Knowledge Web NoE (FP6-507482), 2007.

\bibitem[\protect\citeauthoryear{Euzenat}{2004}]{euzenat:2004}
J{\'e}r{\^o}me Euzenat.
\newblock An {API} for ontology alignment.
\newblock In Frank {van Harmelen}, Sheila McIlraith, and Dimitri Plexousakis,
  editors, {\em The Semantic Web - ISWC 2004: Third International Semantic Web
  Conference,Hiroshima, Japan, November 7-11, 2004. Proceedings}, volume 3298,
  pages 698--712. Springer, 2004.

\bibitem[\protect\citeauthoryear{Fellegi and Sunter}{1969}]{fellegi:1969}
Ivan Fellegi and Alan Sunter.
\newblock A theory for record linkage.
\newblock {\em Journal of the American Statistical Association},
  64(328):1183--1210, 1969.

\bibitem[\protect\citeauthoryear{Hassanzadeh \bgroup \em et al.\egroup
  }{2009}]{hassanzadeh:2009}
Oktie Hassanzadeh, Lipyeow Lim, Anastasios Kementsietsidis, and Min Wang.
\newblock A declarative framework for semantic link discovery over relational
  data.
\newblock In {\em WWW '09: Proceedings of the 18th international conference on
  World wide web}, pages 1101--1102, New York, NY, USA, 2009. ACM.

\bibitem[\protect\citeauthoryear{Heath and Bizer}{2011}]{heath2011a}
Tom Heath and Christian Bizer.
\newblock {\em Linked Data: Evolving the Web into a Global Data Space},
  volume~1 of {\em Synthesis Lectures on the Semantic Web: Theory and
  Technology}.
\newblock Morgan \& Claypool, 1 edition, 2011.

\bibitem[\protect\citeauthoryear{Hogan \bgroup \em et al.\egroup
  }{2007}]{hogan:2007}
Aidan Hogan, Andreas Harth, and Stefan Decker.
\newblock Performing object consolidation on the semantic web data graph.
\newblock In {\em In Proceedings of 1st I3: Identity, Identifiers,
  Identification Workshop}, 2007.

\bibitem[\protect\citeauthoryear{Jaffri \bgroup \em et al.\egroup
  }{2008}]{jaffri:2008b}
Afraz Jaffri, Hugh Glaser, and Ian Millard.
\newblock Managing uri synonymity to enable consistent reference on the
  semantic web.
\newblock In {\em IRSW2008 - Identity and Reference on the Semantic Web 2008 at
  ESWC}, 2008.

\bibitem[\protect\citeauthoryear{Klein}{2001}]{klein:2001}
Michael Klein.
\newblock Combining and relating ontologies: an analysis of problems and
  solutions.
\newblock In {\em Workshop on Ontologies and Information Sharing}, 2001.

\bibitem[\protect\citeauthoryear{Nikolov \bgroup \em et al.\egroup
  }{2008}]{nikolov:2008}
Andriy Nikolov, Victoria Uren, Enrico Motta, and Anne {de Roeck}.
\newblock Handling instance coreferencing in the knofuss architecture.
\newblock In {\em Proceedings of the workshop: Identity and Reference on the
  Semantic Web at 5th European Semantic Web Conference (ESWC 2008)}, 2008.

\bibitem[\protect\citeauthoryear{Noy and Musen}{2000}]{noy:2000}
Natalya~Fridman Noy and Mark~A. Musen.
\newblock Prompt: Algorithm and tool for automated ontology merging and
  alignment.
\newblock In {\em Proceedings of the Seventeenth National Conference on
  Artificial Intelligence and Twelfth Conference on Innovative Applications of
  Artificial Intelligence}, pages 450--455. AAAI Press / The MIT Press, 2000.

\bibitem[\protect\citeauthoryear{Rahm and Bernstein}{2001}]{rahm:2001}
Erhard Rahm and Philip Bernstein.
\newblock A survey of approaches to automatic schema matching.
\newblock {\em VLDB Journal: Very Large Data Bases}, 10(4):334--350, 2001.

\bibitem[\protect\citeauthoryear{Raimond \bgroup \em et al.\egroup
  }{2007}]{raimond:2007}
Yves Raimond, Samer Abdallah, Mark Sandler, and Frederick Giasson.
\newblock The music ontology.
\newblock In {\em Proceedings of the International Conference on Music
  Information Retrieval}, 2007.

\bibitem[\protect\citeauthoryear{Raimond \bgroup \em et al.\egroup
  }{2008}]{raimond:2008}
Yves Raimond, Christopher Sutton, and Mark Sandler.
\newblock Automtic interlinking of music datasets on the semantic web.
\newblock In {\em Proceedings of the Linking Data On the Web workshop at
  WWW'2008}, 2008.

\bibitem[\protect\citeauthoryear{Sahoo \bgroup \em et al.\egroup
  }{2009}]{sahoo2009a}
Satya Sahoo, Wolfgang Halb, Sebastian Hellmann, Kingsley Idehen, Ted Thibodeau,
  S{\"o}ren Auer, Juan Sequeda, and Ahmet Ezzat.
\newblock A survey of current approaches for mapping relational databases to
  rdf.
\newblock Report, W3C RDB2RDF incubator group, 2009.

\bibitem[\protect\citeauthoryear{Sa\"{i}s \bgroup \em et al.\egroup
  }{2008}]{sais:2008}
Fatia Sa\"{i}s, Nathalie Pernelle, and Marie-Christine Rousset.
\newblock Combining a logical and a numerical method for data reconciliation.
\newblock {\em Journal of Data Semantics}, 12, 2008.

\bibitem[\protect\citeauthoryear{Sauermann and Cyganiak}{2008}]{sauermann:2008}
Leo Sauermann and Richard Cyganiak.
\newblock Cool {URIs} for the semantic web.
\newblock {W3C} note, W3C, March 2008.
\newblock http://www.w3.org/TR/2008/NOTE-cooluris-20080331/.

\bibitem[\protect\citeauthoryear{{Scharffe} and
  {Euzenat}}{2010}]{scharffe2010a}
{François} {Scharffe} and {J\'er\^ome} {Euzenat}.
\newblock M\'ethodes et outils pour lier le web des donn\'ees.
\newblock In {\em Actes 17e conf\'erence AFIA-AFRIF sur reconnaissance des
  formes et intelligence artificielle (RFIA), Caen (FR)}, pages 678--685, 2010.

\bibitem[\protect\citeauthoryear{Scharffe \bgroup \em et al.\egroup
  }{2009}]{scharffe:2009}
Fran\c{c}ois Scharffe, Yanbin Liu, and Chunguang Zhou.
\newblock {RDF-AI}: an architecture for {RDF} datasets matching, fusion and
  interlink.
\newblock In {\em Workshop on Identity and Reference in Knowledge
  Representation, IJCAI 2009}, 2009.

\bibitem[\protect\citeauthoryear{Visser \bgroup \em et al.\egroup
  }{1997}]{visser:1997}
Pepjijn R.~S. Visser, Dean~M. Jones, T.~J.~M. Bench-Capon, and M.~J.~R. Shave.
\newblock An analysis of ontological mismatches: Heterogeneity versus
  interoperability.
\newblock In {\em {AAAI} 1997 Spring Symposium on Ontological Engineering},
  Stanford, USA, 1997.

\bibitem[\protect\citeauthoryear{V{\"o}lkel \bgroup \em et al.\egroup
  }{2006}]{volkel:2006}
Max V{\"o}lkel, Markus Kr{\"o}tzsch, Denny Vrandecic, Heiko Haller, and Rudi
  Studer.
\newblock Semantic wikipedia.
\newblock In {\em WWW}, pages 585--594, 2006.

\bibitem[\protect\citeauthoryear{Volz \bgroup \em et al.\egroup
  }{2009}]{volz2009a}
Julius Volz, Christian Bizer, and Martin Gaedke.
\newblock Web of data link maintenance protocol.
\newblock Protocol specification, Frei Universit{\"a}t Berlin, 2009.

\bibitem[\protect\citeauthoryear{Winkler}{2006}]{winkler:2006}
William Winkler.
\newblock Overview of record linkage and current research directions.
\newblock Technical Report 2006-2, Statistical Research Division. U.S. Census
  Bureau, 2006.

\end{thebibliography}

\newpage
\tableofcontents

\end{document}